%% file: main.tex
\documentclass[10pt,twocolumn,letterpaper]{article}

\usepackage[pagenumbers]{cvpr}

\input{preamble}

\definecolor{cvprblue}{rgb}{0.21,0.49,0.74}
\usepackage[pagebackref,breaklinks,colorlinks,allcolors=cvprblue]{hyperref}

\usepackage{multirow}
\usepackage{enumitem}
\usepackage{makecell}
\usepackage{adjustbox}

\def\confName{CVPR}
\def\confYear{2027}

\title{Thinking in Streams: Proactive Spatial Reasoning with Vision-Language Models}

\author{
Junyoung Hong \quad Minsu Kim \quad Jaehyun Kim \quad Jeonghwan Cho \quad Jiwoo Lee \quad Seon Joo Kim\\[1.0em]
Yonsei University
}

\begin{document}

\twocolumn[{%
    \renewcommand\twocolumn[1][]{#1}%
    \maketitle
    \input{main/partial/fig_teaser}
}]

\input{macro}
\input{main/section/abstract}
\input{main/section/introduction}
\input{main/partial/fig_task_overview}
\input{main/section/related_work}
\input{main/section/prospace}

\input{main/section/experiments}
\input{main/section/conclusion}

\section*{LLM Usage Statement}
We used ChatGPT (OpenAI) for manuscript editing and teaser image generation, and Claude (Anthropic) for code development assistance.
All scientific claims and experimental results were verified by the authors.

{
    \small
    \bibliographystyle{ieeenat_fullname}
    \bibliography{main}
}

\end{document}

%% file: preamble.tex
\usepackage{booktabs}
\usepackage{multirow}
\usepackage{graphicx}
\usepackage{makecell}
\usepackage{xcolor}
\usepackage{colortbl}
\usepackage{dsfont}
\usepackage{float}

%% file: main/partial/fig_teaser.tex
\vspace{-9pt}
\noindent
\begin{minipage}{\textwidth}
    \centering
    \includegraphics[width=\textwidth]{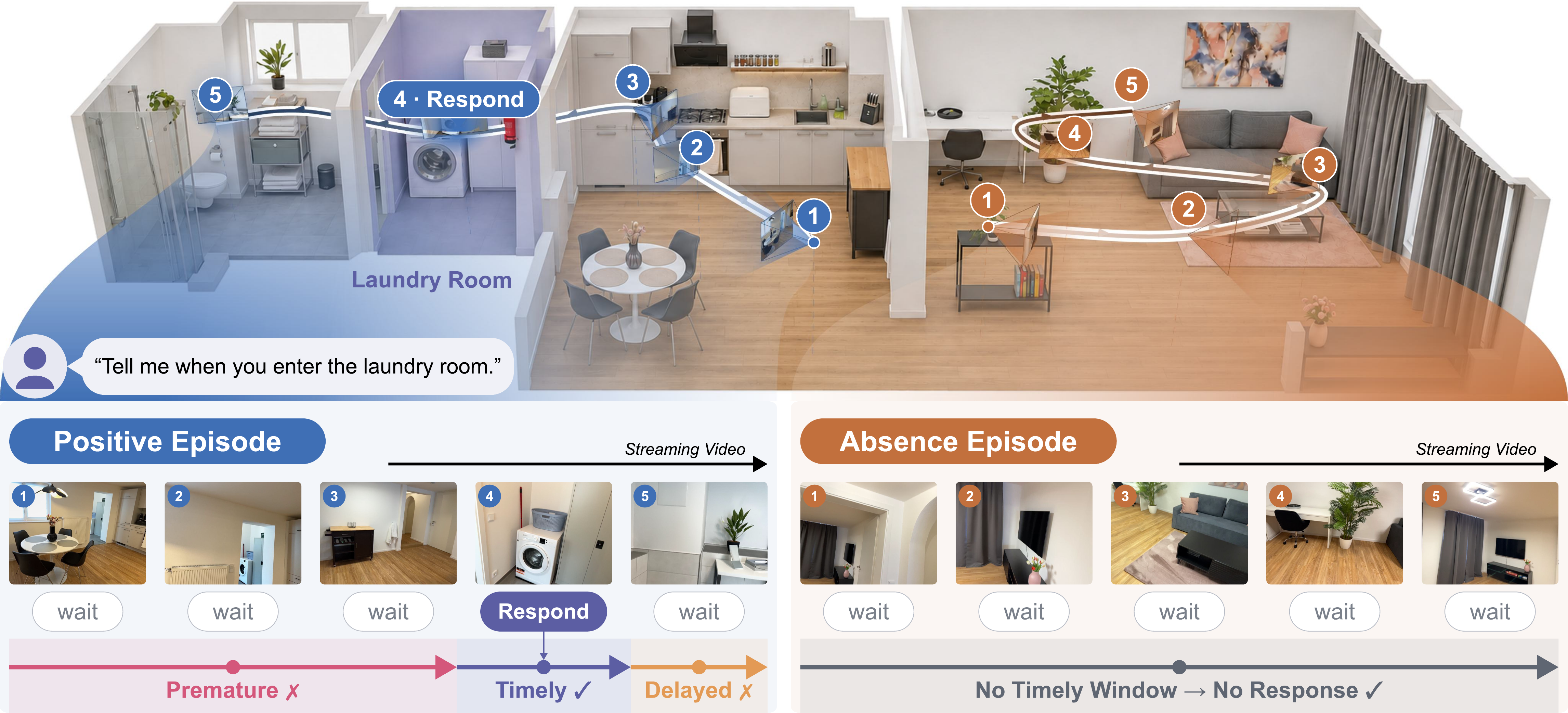}
    \captionsetup{hypcap=false}
    \captionof{figure}{
        \textbf{ProSpace} brings spatial reasoning to streaming video in embodied environments, where models \textit{think in streams} to decide when to respond to spatial requests. This requires maintaining and updating spatial knowledge across changing viewpoints.
    }
    \label{fig:teaser}
\end{minipage}
\par\vspace{10pt}

%% file: macro.tex
\newcommand{\minsu}[1]{{\color{orange}#1}}
\newcommand{\minsured}[1]{{\color{red}#1}}
\newcommand{\junyoung}[1]{{\color{blue}#1}}

%% file: main/section/abstract.tex
\begin{abstract}
Recognizing when a requested spatial condition holds in streaming video can require maintaining and updating spatial knowledge across changing viewpoints.
We study this capability through proactive spatial reasoning, where a model receives a query in advance and determines whether and when to respond as the stream unfolds.
We introduce ProSpace-Bench, comprising nine tasks across object, camera, and scene reasoning, and ProSpace-Train for supervised training.
To support spatial decisions, we propose ProSpace-VLM, which injects explicit geometric encodings estimated causally from incoming frames into intermediate VLM layers, using the existing KV cache as spatial memory and spatial token selection for efficient streaming.
Experiments reveal a substantial gap between VLMs and humans, while ProSpace-VLM improves timely responses and correct abstention over fine-tuned baselines.
Analyses examine how ProSpace-VLM represents previously observed spatial information across changing viewpoints.
\end{abstract}

%% file: main/section/introduction.tex
\section{Introduction}
\label{sec:introduction}

When we move through a room, we see only part of it at a time, yet come to understand how the whole room is laid out.
{Humans do so by building and updating a mental model of space~\cite{shepard1986mentalimages,gardner2011framesofmind}, an internal representation central to spatial embodied intelligence~\cite{yang2025thinkinginspace,wu2026spatialmllm,chen2024spatialvlm,wang2025mindcube, zhang2026theoryofspace}.
As vision-language models (VLMs) advance, a key goal is to enable them to build, update, and use such representations while processing a continuous stream of visual observations.}

Research toward this goal has examined how VLMs represent and reason about space~\cite{yang2025thinkinginspace,wu2026spatialmllm} and has expanded to online settings, where models accumulate and maintain spatial knowledge as observations arrive~\cite{lin2026ostbench,li2026ovosbench,xie2026streamspatial,wang2026keepitinmind,yu2026stream3dvlm}.
Existing evaluations primarily assess what models can infer from observations provided up to predefined query times.
We explore a different perspective on spatial mental modeling by examining how models decide when to respond as a video stream unfolds.

{To this end, we formulate \emph{proactive spatial reasoning} in streaming video.
Given a query in advance, a model responds when its spatial condition is satisfied in an incoming video stream and remains silent if the condition is never met.
For example, a user might ask an assistant, “Remind me to pick up my bag when I return to this room.”
A key observation is that deciding {whether and when to respond} as a stream unfolds can require models to maintain and update spatial knowledge across changing viewpoints.}

Building on this formulation, we introduce \textbf{ProSpace}, a framework for evaluating and developing proactive spatial reasoning in streaming video.
Its benchmark, \textbf{ProSpace-Bench}, comprises nine tasks across the object, camera, and scene levels.
These tasks examine how objects relate to the observer,  how the camera moves over time, and where the observer is in the global environment.

By jointly assessing spatial understanding, response timing, and abstention, ProSpace-Bench probes whether models can use relevant spatial knowledge at the appropriate moment as a stream unfolds.
For supervised fine-tuning, we introduce ProSpace-Train, comprising 100K streams with supervision for response timing and abstention.

To support continual spatial decisions, we propose \textbf{ProSpace-VLM}, a geometry-aware streaming VLM that uses the existing KV cache as spatial memory.
To write spatial information into this cache, we align causally estimated 3D points and camera poses~\cite{wang2025cut3r} with visual tokens and inject their encodings into intermediate VLM layers.
As frames arrive, spatially grounded key--value representations accumulate in the cache, and the model reads them through causal attention while processing each new frame to decide whether to respond.
This alignment also associates cached tokens with 3D positions, enabling voxel-based token selection for efficient streaming inference.

ProSpace-Bench reveals a substantial gap between VLMs and humans in proactive spatial reasoning.
ProSpace-VLM narrows this gap, improving both timely response and correct abstention over fine-tuned general-purpose and geometry-aware baselines.
Further analyses suggest that ProSpace-VLM retains spatial information from past observations, with attention visualization in 3D space and probes examining whether it maintains beliefs about objects' relative directions after they leave view.

Our contributions are (i) the formulation of \emph{proactive spatial reasoning}, a task assessing how models use spatial information over time to decide whether and when to respond; (ii) \textbf{ProSpace-Bench}, a nine-task benchmark evaluating spatial understanding, response timing, and abstention; and (iii) \textbf{ProSpace-VLM}, a geometry-aware streaming VLM injecting explicit 3D geometry into intermediate layers and accumulating spatially grounded visual representations in its existing KV cache.

%% file: main/partial/fig_task_overview.tex
\begin{figure*}[t]
    \centering
    \includegraphics[width=\textwidth]{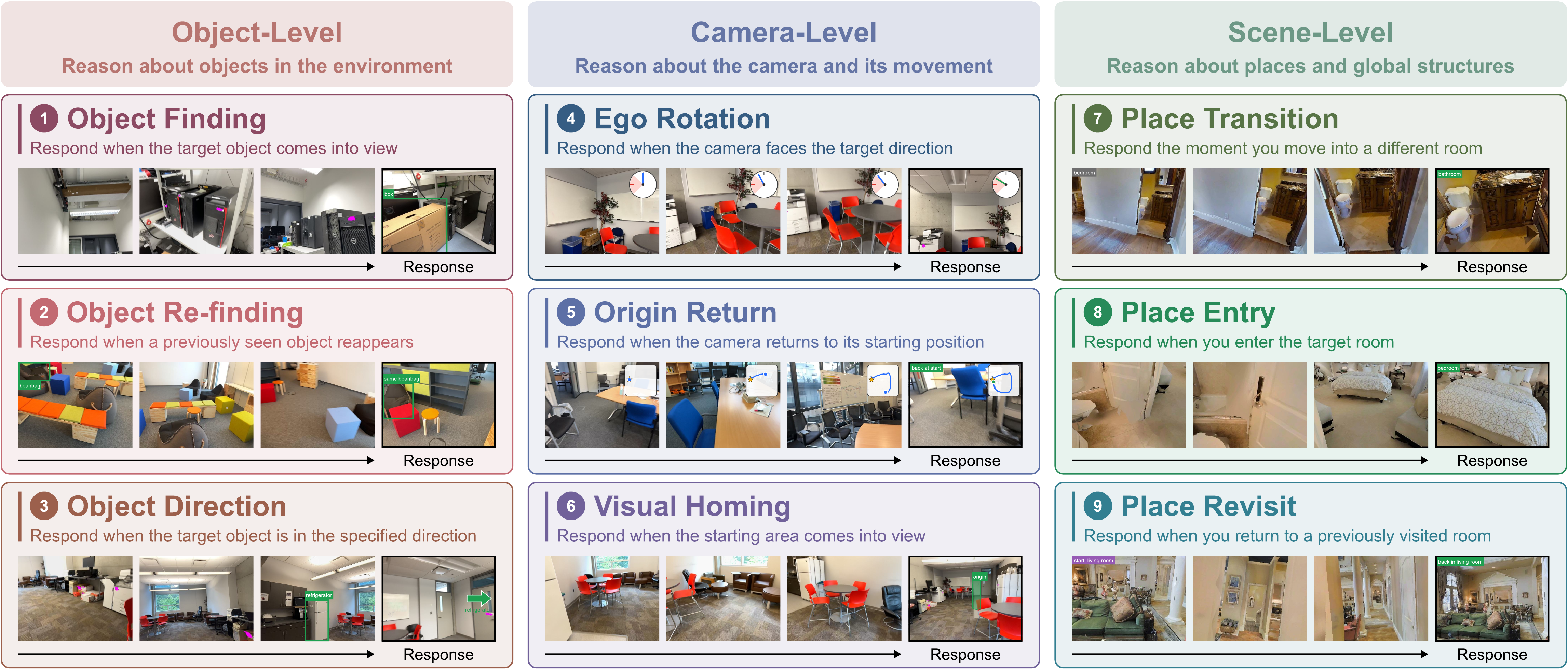}
    \caption{\textbf{Task overview of ProSpace-Bench.}
    Bounding boxes mark target objects in object-level tasks.
    The blue needle in \textit{Ego Rotation} indicates the camera's viewing direction; the star in \textit{Origin Return} and the cylinder in \textit{Visual Homing} mark the starting position.
    }
    \label{fig:task_overview}
    \vspace{-4mm}
\end{figure*}

%% file: main/section/related_work.tex
\section{Related Work}
\label{sec:related_work}

\subsection{Visual-Spatial Intelligence}
Visual-spatial intelligence, a core aspect of human cognition, encompasses the ability to perceive, represent, and mentally manipulate spatial structures and relations~\cite{shepard1986mentalimages,gardner2011framesofmind}.
Its importance for embodied perception and interaction motivates efforts to evaluate and improve this capability in vision-language models.
Image-based benchmarks assess spatial relations, depth ordering, size, and distance~\cite{tong2024cambrian1,liao2024spatialprompt,du2024embspatialbench,ma20253dsrbench}, while multi-view and video benchmarks test integration across views and time, perspective taking, and spatial memory~\cite{yang2025thinkinginspace,wang2025site,wang2025mindcube,yang2026mmsibench,li2026viewspatialbench}.
Methods improve these capabilities through spatially grounded supervision, depth-aware representations, and spatial instruction tuning~\cite{chen2024spatialvlm,cheng2024spatialrgpt,cai2025spatialbot,ma2025spatialllm,yang2026vst,cai2026sensenovasi}, or explicit 3D geometry and learned geometric priors~\cite{zhu2024llava3d,fan2026vlm3r,zheng2026vgllm}.

Beyond fixed images and video clips, recent work studies spatial intelligence in online scenarios, where observations arrive incrementally and future frames remain unavailable. %
Benchmarks in this online setting assess reasoning during scene exploration~\cite{lin2026ostbench}, egocentric perception and allocentric mapping~\cite{li2026ovosbench}, and reasoning about past, present, and future spatial states~\cite{xie2026streamspatial}.
To maintain and update spatial context, recent approaches use structured spatial memory~\cite{wang2026keepitinmind,liu2026onlinesi}, adaptive fast weights~\cite{liu2026spatialttt}, and prediction-guided memory selection~\cite{yang2026cambrians}.
Despite this progress, these evaluations largely require models to answer at predefined times.
Stream3D-Bench~\cite{yu2026stream3dvlm} studies a related setting, where models receive spatial requests in advance and respond later in the stream.
However, most of its questions specify a waiting duration.
A detailed comparison is provided in the supplementary material.

Instead, ProSpace-Bench requires models to independently determine when spatial conditions are satisfied, including conditions that require relating past observations to the observer's current state.
To support this capability, we design ProSpace-VLM for efficient proactive spatial reasoning in streaming video.
It causally integrates explicit geometric encodings rather than latent spatial features~\cite{fan2026vlm3r,yu2026stream3dvlm,zheng2026vgllm,zhao_2026_spacemind,wu2026spatialmllm} to enable the existing KV cache to serve as spatially aware streaming memory.

\subsection{Proactive Response in Streaming Video}
Proactive response in streaming video requires models to monitor incoming observations and autonomously decide when to respond.
Recent methods address this task through learned response policies and decision mechanisms~\cite{wang2025mmduet,qian2025dispider,wang2026mmduet2}.
Existing benchmarks evaluate responses as events unfold or sufficient visual evidence becomes available~\cite{niu2025ovobench,wang2025proactivevideoqa,azad2026streamready,chatterjee2026dontpause}.
While these benchmarks evaluate proactive responses across general video understanding tasks, our benchmark focuses on reasoning about spatial conditions.

%% file: main/section/prospace.tex
\section{ProSpace}
\label{sec:prospace}

\subsection{Task Formulation}
\label{sec:task_formulation}
An episode consists of a spatial request $q$ and a video stream $V=(v_0,\ldots,v_T)$.
The model receives the spatial request in advance and then incrementally processes the video stream as frames arrive.
At each time step, the model decides whether to wait or respond based only on observations received so far.
Let $\hat{t}\in\{0,\ldots,T\}\cup\{\infty\}$ denote the first response time, where $\hat{t}=\infty$ indicates no response.
We distinguish between \emph{positive} episodes, in which the requested spatial condition occurs once, and \emph{absence} episodes, in which it never occurs.
For a positive episode, we define a valid response window $W=[t_s,t_e]$ as the temporal interval during which the requested spatial condition is satisfied.
A response is \emph{premature} if $\hat{t}<t_s$, \emph{timely} if $t_s\leq\hat{t}\leq t_e$, and \emph{delayed} if $t_e<\hat{t}<\infty$.
A positive episode with $\hat{t}=\infty$ is considered a \emph{miss}.
For an \emph{absence} episode, the requested spatial condition never occurs and $W=\varnothing$.
A model achieves \emph{correct abstention} if $\hat{t}=\infty$, while any response is considered a \emph{false alarm}.

\input{main/section/prospace_bench}
\input{main/section/prospace_vlm}

%% file: main/section/prospace_bench.tex
\subsection{ProSpace-Bench}
\label{sec:prospace_bench}

\subsubsection{Task Overview}
\label{sec:task_overview}
To evaluate proactive spatial reasoning in streaming video, ProSpace introduces nine tasks organized into three groups that probe complementary spatial reasoning capabilities required to determine when a request is satisfied (Fig.~\ref{fig:task_overview}).

\vspace{-0.5em}
\paragraph{Object.}
Object-level tasks focus on individual objects and their spatial relationship to the observer.
\textit{Object Finding} tests the basic ability to detect when a target object first comes into view.
\textit{Object Re-finding} evaluates recognition of a target object specified in the first frame when it reappears later.
\textit{Object Direction} requires determining when a previously observed object lies in a specified direction relative to the camera, even when it is no longer visible.

\vspace{-0.5em}
\paragraph{Camera.}
Camera-level tasks require reasoning about spatial conditions defined by the camera's own motion.
\textit{Ego Rotation} requires detecting when the camera faces a specified direction relative to its initial pose.
\textit{Origin Return} requires determining when the camera returns to its starting position, regardless of viewing direction.
\textit{Visual Homing} requires recognizing when the starting region becomes visible again without physically returning to it.

\vspace{-0.5em}
\paragraph{Scene.}
Scene-level tasks consider the broader spatial organization of the environment and the observer's location within it.
\textit{Place Transition} requires detecting the moment of transition between rooms.
\textit{Place Entry} requires responding while inside a specified room.
\textit{Place Revisit} requires recognizing a return to a previously visited room.

\input{main/partial/fig_architecture}

\subsubsection{Benchmark Construction}
\label{sec:benchmark_construction}

\paragraph{Sources and splits.}
We construct the object-level and camera-level tasks using the official validation splits of ScanNet~\cite{dai2017scannet}, ScanNet++~\cite{yeshwanth2023scannetpp}, and CA-1M~\cite{lazarow2025cubify}, which provide indoor video sequences with camera poses and object annotations.
For the scene-level tasks, we use Matterport3D~\cite{chang2017matterport3d} validation and test houses with room-level annotations. Overall, ProSpace-Bench comprises 3,636 episodes across 242 real indoor scenes and 18 houses, totaling about 61 hours of video, sampled at 1 fps.

\paragraph{Episode construction.}

We derive per-frame metadata from ground-truth camera poses and scene meshes, including object visibility, relative bearings, camera motion, and room membership, and apply task-specific criteria to determine valid response windows.
For positive episodes, we retain a single contiguous window with sufficient context before, during, and after it to evaluate premature, timely, and delayed responses.
Absence episodes follow the same request prompt as positive episodes, but the requested condition is never satisfied.
To avoid trivial absence episodes, we include task-specific distractors, such as another object of the same class or an approach to the target room without entering it.

For quality control, we check for confusable object labels, verify object visibility, and filter out episodes with discontinuous camera poses.
For scene-level tasks, we also inspect rendered videos for visual defects.
These checks help ensure that each episode is clear and answerable for human observers.
We further review pilot episodes for each task through human evaluation to identify remaining annotation issues.
Details about ProSpace-Bench construction are provided in the supplementary material.

\paragraph{ProSpace-Train.}
To support supervised fine-tuning, we construct ProSpace-Train, comprising 100K positive and absence streams built from the official training splits of the benchmark datasets with relaxed episode criteria, sharing no scene with ProSpace-Bench.
We cap the number of samples from frequent tasks to reduce imbalance in the training distribution.
To prevent models from exploiting elapsed time as a shortcut, we vary both the request time and the start of the response window across streams.
Further details are provided in the supplementary material.

%% file: main/partial/fig_architecture.tex
\begin{figure}[t]
    \centering
    \includegraphics[width=\columnwidth]{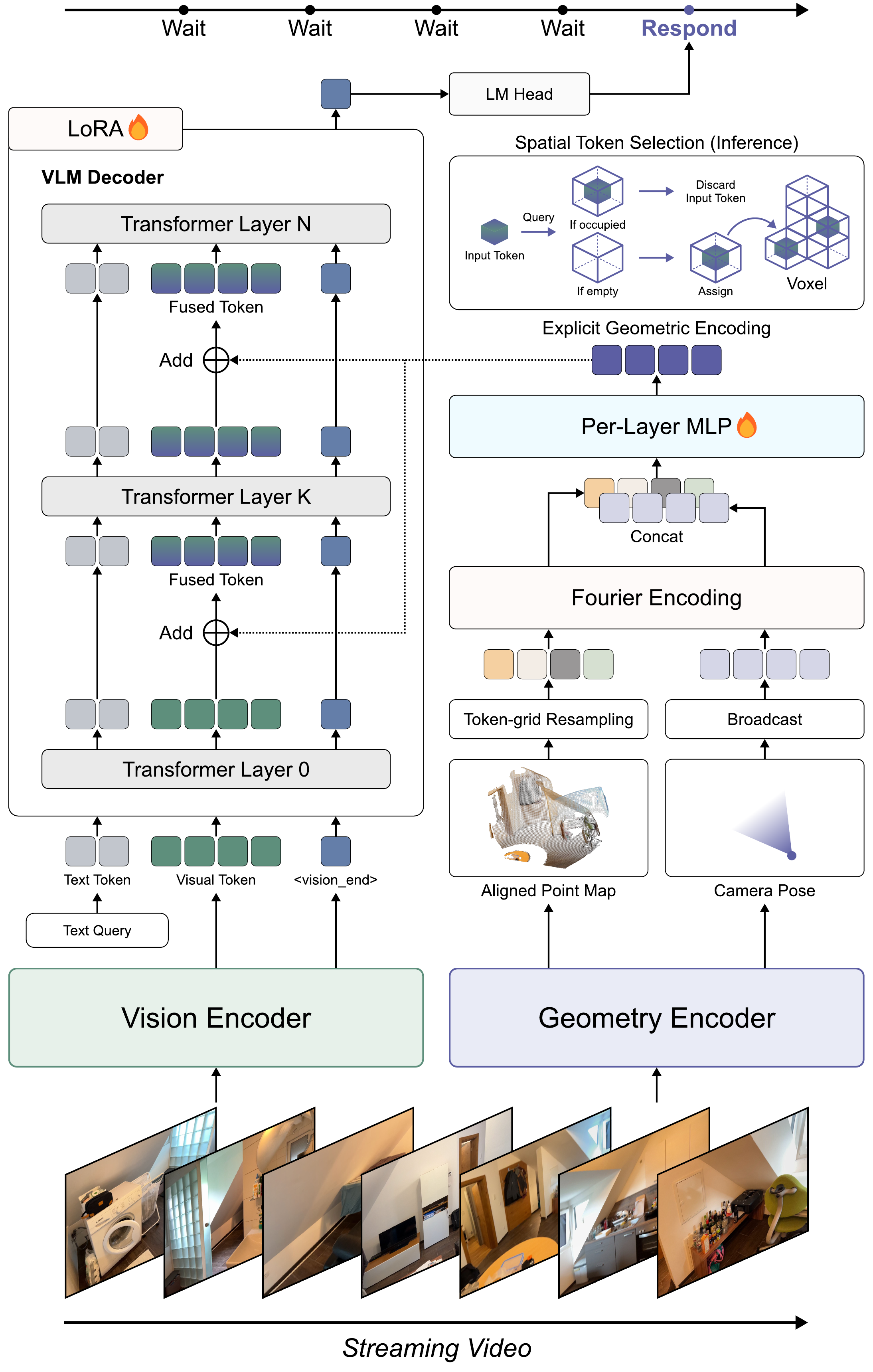}
    \caption{\textbf{Architecture of ProSpace-VLM}.
    A causal geometry encoder estimates 3D point maps and camera poses from incoming frames. The resulting geometric encodings are injected into visual tokens.
    The model reads the cache through causal attention to assess the current spatial state and decide when to respond.
    }
    \label{fig:architecture}
    \vspace{-4mm}
\end{figure}

%% file: main/section/prospace_vlm.tex
\subsection{ProSpace-VLM}
\label{sec:prospace_vlm}

We introduce ProSpace-VLM, a geometry-aware streaming VLM shown in Fig.~\ref{fig:architecture}.
We use the existing KV cache as spatial memory, relying on its native write and read operations for effective proactive spatial reasoning.

\subsubsection{Encoding}
\label{sec:encoding}

ProSpace-VLM receives a spatial request in advance and processes the video stream causally, one frame at a time.
For each incoming frame, a vision encoder extracts visual tokens, while the geometry encoder CUT3R~\cite{wang2025cut3r} estimates a 3D point map and camera pose.
While prior approaches primarily use latent spatial features to incorporate 3D information into VLMs~\cite{fan2026vlm3r,yu2026stream3dvlm,zheng2026vgllm,zhao_2026_spacemind,wu2026spatialmllm}, we use explicit point maps and camera poses to align geometry with visual tokens.
To align them, we resample the point map to the visual token grid, associating each visual token $j$ with a 3D position $\mathbf{x}_{t,j}$.
The camera pose is represented by its center $\mathbf{c}_t$ and rotation $\mathbf{R}_t$.
We construct the geometric encoding as
\begin{equation}
\mathbf{g}_{t,j} =
\left[
\phi(\mathbf{x}_{t,j}),
\psi(\mathbf{c}_t,\mathbf{R}_t)
\right],
\end{equation}
where $\phi$ encodes the 3D position using Fourier features~\cite{mildenhall2020nerf}, and $\psi$ encodes the camera pose using the same features together with a continuous 6D rotation representation~\cite{zhou2019continuity}.
The geometric encoding combines the point and camera pose encoding through concatenation.

\subsubsection{KV Cache as Spatial Memory}
\label{sec:spatial_memory}

Since the VLMs maintain a KV cache at each layer, we inject geometric encoding at multiple layers.
This is implemented by element-wise addition to the visual tokens:
\begin{equation}
\widetilde{\mathbf{h}}_{t,j}^{(\ell)}
=
\mathbf{h}_{t,j}^{(\ell)}
+
f_\ell(\mathbf{g}_{t,j}),
\qquad \ell \in \mathcal{L},
\end{equation}
where $\mathbf{h}_{t,j}^{(\ell)}$ denotes the representation of visual token $j$ from frame $t$ at layer $\ell$, and $f_\ell$ is a layer-specific lightweight MLP.
We select early-to-middle layers to integrate geometric encoding while allowing later layers to perform high-level reasoning.
Through this process, the geometry-aware key--value representations accumulate in the KV cache as new frames arrive.
The model reads from this spatial memory using standard causal attention, relating the current observation to the accumulated spatial history to decide whether to respond.

\subsubsection{Spatial Token Selection}
\label{sec:spatial_cache}
With each cached visual token associated with a 3D point, we can apply voxel-based token selection to the KV cache for efficient streaming inference.
We retain recent frames in full and select tokens from older frames by keeping only the earliest cached token in each voxel, increasing the voxel size as needed to meet the token budget.
Our spatial memory is anchored in a common coordinate frame, enabling token selection across the stream.

\subsubsection{Streaming Response Decision}
\label{sec:streaming_response_decision}

To make a decision as each frame arrives, ProSpace-VLM uses next-token prediction to choose between \(\texttt{wait}\) and \(\texttt{respond}\), following the streaming response-control formulation of Stream3D-VLM~\cite{yu2026stream3dvlm}.
We train the geometric projection MLPs and adapt the language model using LoRA~\cite{hu2021lora} with cross-entropy loss over the decision tokens:
\begin{equation}
\mathcal{L}_{\mathrm{resp}}
=
-\sum_{t\in\mathcal{S}}
\log P_\theta(y_t\mid q,v_{0:t},y_{<t}),
\end{equation}
where $y_t$ denotes the target decision and $\mathcal{S}$ is the set of supervised frame indices.

At inference, the model reads $P_\theta(\texttt{respond} \mid q, v_{0:t})$, the softmax over the two decision tokens, after each frame.
The response time is defined as
\begin{equation}
\hat{t} = \min\{\, t \ge t_q : P_\theta(\texttt{respond} \mid q, v_{0:t}) \ge \tau \,\},
\end{equation}
where $t_q$ is the request time and $\tau$ is the response threshold.
If no response is triggered, the model remains silent.

%% file: main/section/experiments.tex
\section{Experiments}
\label{sec:experiments}

\input{main/table/table_prospace_bench}

\subsection{Experimental Setup}
\label{sec:experimental_setup}

\paragraph{Models and Baselines.}
We evaluate proprietary~\cite{hurst2024gpt4o,google2026gemini31pro,anthropic2026claudeopus55}, general-purpose~\cite{qwen2025qwen25vl,bai2025qwen3}, spatially specialized~\cite{cosmosreason2_32b,cai2026sensenovasi,yang2026cambrians,fan2026vlm3r}, and streaming VLMs~\cite{yao2025timechatonline,chen2025livecc,qian2025dispider,liu2026spatialttt,yu2026stream3dvlm} on ProSpace-Bench for zero-shot evaluations.
As reference baselines, \textit{Random} selects a response time uniformly within each episode, while \textit{Fixed-Time Oracle} selects the best fixed response time for each task using benchmark annotations; both are evaluated on positive episodes only.
\textit{Blind VLM} evaluates Claude Opus 5.5~\cite{anthropic2026claudeopus55} under the same protocol without any video frames.
We also evaluate human performance on 900 episodes, with 100 per task, under the same streaming setting as the models.

\paragraph{Training Protocol.}
We fine-tune all models~\cite{bai2025qwen3,fan2026vlm3r,liu2026spatialttt,chen2024videollmonline,yu2026stream3dvlm} on ProSpace-Train for one epoch using cross-entropy loss, adapting response-window annotations to each inference interface.
Non-streaming VLMs use video prefixes labeled \texttt{Yes} when the final frame falls within the response window and \texttt{No} otherwise.
Absence prefixes come from both positive and absence streams, including hard negatives immediately before the window.
Streaming VLMs receive \texttt{wait} supervision before the window and \texttt{respond} within it, with \texttt{wait} throughout absence streams.

\paragraph{Evaluation Protocol.}
All models process observations causally, without access to future frames.
For non-streaming VLMs, we use prefix polling inspired by OVO-Bench~\cite{niu2025ovobench}: at each time step, the model receives the observed video prefix and predicts whether the requested condition holds.
Streaming VLMs receive the request once and process incoming frames using their native response mechanisms.
In both cases, the first ``yes'' response determines the response time $\hat{t}$.

\input{main/partial/fig_analysis_attention}

\paragraph{Metric Design.}
We measure timely responses on positive episodes and correct abstention on absence episodes:
\begin{align}
\mathrm{Timely\ Response}
&= \frac{1}{|\mathcal{P}|}
\sum_{i\in\mathcal{P}}
\mathds{1}\left[t_{s,i} \leq \hat{t}_i \leq t_{e,i}\right],\\
\mathrm{Correct\ Abstention}
&= \frac{1}{|\mathcal{N}|}
\sum_{j\in\mathcal{N}}
\mathds{1}\left[\hat{t}_j = \infty\right].
\end{align}
Here, $\mathcal{P}$ and $\mathcal{N}$ denote the sets of positive and absence episodes.
We report the average of each metric across the nine tasks.

\input{main/partial/fig_analysis_bench_breakdown}
\input{main/partial/fig_analysis_failure_case}

\input{main/table/table_ablation}
\input{main/table/table_spatial_bench}

\subsection{Main Results}
\label{sec:main_results}

\paragraph{Results on ProSpace-Bench.}
As shown in \cref{tab:prospace_bench}, existing VLMs fall substantially short of human performance on proactive spatial reasoning, with several achieving timely response rates near or below the random-timing baseline.
Models perform better on tasks such as \textit{Object Finding}, where the requested condition can be assessed through single-frame object recognition, but struggle on tasks that require relating observations across changing viewpoints.
Some streaming VLMs achieve high correct abstention but no timely responses on certain tasks.
High correct abstention alone does not indicate good performance if the model fails to respond when needed.
\cref{fig:analysis_bench_breakdown} further shows that zero-shot failures are dominated by premature responses or missed responses, whereas delayed responses are rare across all groups.
\cref{fig:analysis_failure_case} illustrates failure cases of a zero-shot model.
To inspect the reasoning behind these errors, we re-query GPT-4o~\cite{hurst2024gpt4o} at its response time with the same frames and question, replacing the one-word answer instruction with a request for step-by-step spatial reasoning.
Supervised fine-tuning on ProSpace-Train substantially improves both timely responses and correct abstention across prefix-polling and streaming models.

ProSpace-VLM outperforms fine-tuned spatial and online video baselines, even on challenging tasks that require continuous integration of past observations.
These results suggest that injecting explicit geometric encodings into intermediate VLM layers enables the KV cache to serve as spatial memory for proactive spatial reasoning, offering an alternative to fast-weight memory~\cite{liu2026spatialttt} and input-level fusion of latent spatial features~\cite{fan2026vlm3r,yu2026stream3dvlm}.

\subsection{Analysis}
\label{sec:analysis}

\paragraph{Ablation study.}
We ablate individual components in the final entry of \cref{tab:prospace_bench} and in \cref{tab:ablation}, examining the geometry encoder~\cite{wang2025cut3r,zhuo2026streamvggt}, geometric inputs, and single-layer injection at the input.
Our model achieves comparable results with metric-scale CUT3R~\cite{wang2025cut3r} and non-metric StreamVGGT~\cite{zhuo2026streamvggt}, outperforming all baselines on both metrics.
Injecting geometry into intermediate layers further improves performance over input-level fusion, suggesting that it helps deeper VLM layers access spatial history.

\paragraph{Generalization to spatial benchmarks.}
Our model is fine-tuned exclusively for proactive spatial reasoning, producing respond-or-wait decisions rather than answers to spatial questions.
To assess whether this training transfers beyond the decision setting, we evaluate the fine-tuned models zero-shot on VSI-Bench~\cite{yang2025thinkinginspace} and the online benchmarks OST-Bench~\cite{lin2026ostbench}, UCS-Bench~\cite{wang2026keepitinmind}, and OVO-S-Bench~\cite{li2026ovosbench} (\cref{tab:spatial_bench}).
As ProSpace focuses on proactive spatial reasoning rather than metric-scale estimation, we evaluate the five configurational and spatiotemporal tasks of VSI-Bench~\cite{yang2025thinkinginspace}.
Among the fine-tuned models, our model achieves the highest average score with competitive performance across all four benchmarks.
These results suggest that our geometry-aware design also supports online spatial reasoning over sequential observations, beyond proactive response decisions.

\input{main/table/table_analysis_efficiency}
\input{main/partial/fig_analysis_stream_length}

\paragraph{Analysis of attention visualization.}
We visualize attention from the decision token to cached visual tokens at their associated 3D positions to examine how models access past observations (\cref{fig:analysis_attention}).
In the \textit{Visual Homing} example, the baselines primarily attend to regions around the current viewpoint, while our model consistently attends to the starting region even when the camera is far away.
This suggests that building spatial memory with explicit geometric encoding helps ProSpace-VLM maintain spatial consistency as the viewpoint changes.

\paragraph{Analysis of stream length and absence.}
\cref{fig:analysis_stream_length} evaluates response robustness as the duration of the stream before the response window grows longer.
Our model maintains higher timely response rates on positive episodes and lower per-second response probabilities on absence episodes over time.
These results suggest that our model can sustain spatial mental modeling over extended periods.

\input{main/partial/fig_analysis_objperm}

\paragraph{Analysis of object permanence.}

We analyze spatial mental modeling through object permanence, examining whether models retain objects' spatial locations after they leave view.
For each model, we train a linear probe to decode the target bearing from decision-token states.
As shown in \cref{fig:analysis_objperm}, the directions decoded from our model remain more consistently aligned with the target object as the observer moves (top).
Our model generally maintains lower bearing errors over time after the target leaves view (bottom).
These results suggest that our model's internal representations more accurately encode the evolving spatial relation between a previously observed object and the observer's current viewpoint.

\paragraph{Analysis of efficiency.}
Our approach selects tokens across frames using a shared voxel grid, unlike GAVC~\cite{yu2026stream3dvlm}, which compresses tokens within each frame.
This nearly halves peak memory usage on long streams while maintaining real-time inference at the evaluation rate (\cref{tab:analysis_efficiency}).
Moreover, reusing the existing KV cache as spatial memory enables efficient streaming without a separate retrieval module or additional memory architecture.

Additional experiments and analyses are provided in the supplementary material.

%% file: main/table/table_prospace_bench.tex
\definecolor{navyblue}{HTML}{0071BC}
\definecolor{rankviolet}{HTML}{5C5EA3}

\begin{table*}[t]
    \centering
    \begingroup
    \newcommand{\ts}[2]{#1\,/\,#2}
    \newcommand{\pending}{\textcolor{black!45}{--}}
    \newcommand{\pendingrow}[1]{\textcolor{black!205}{#1} & \pending & \pending & \pending & \pending & \pending & \pending & \pending & \pending & \pending & \pending \\}
    \setlength{\tabcolsep}{3.5pt}
    \renewcommand{\arraystretch}{1.10}
    \setlength{\aboverulesep}{0.3ex}
    \setlength{\belowrulesep}{0.3ex}
    \scriptsize
    \caption{\textbf{Evaluation on ProSpace-Bench.}
    Each cell reports timely response / correct abstention (\%) on positive / absence episodes.
    OF: Object Finding; RF: Object Re-finding; OD: Object Direction;
    ER: Ego Rotation; OR: Origin Return; VH: Visual Homing;
    PT: Place Transition; PE: Place Entry; PR: Place Revisit.
    See the supplementary material for additional results.
    }
    \resizebox{\textwidth}{!}{%
    \begin{tabular}{@{}l c ccc ccc ccc@{}}
        \toprule
        \multirow{2}{*}{Method} &
        \multirow{2}{*}{Average} &
        \multicolumn{3}{c}{Object} &
        \multicolumn{3}{c}{Camera} &
        \multicolumn{3}{c}{Scene} \\
        \cmidrule(lr){3-5}\cmidrule(lr){6-8}\cmidrule(lr){9-11}
        & &
        OF &
        RF &
        OD &
        ER &
        OR &
        VH &
        PT &
        PE &
        PR \\
        \midrule
        \rowcolor{rankviolet!30}[0pt][0pt]
        \multicolumn{11}{@{}l@{}}{\textcolor{black}{{Baseline}}} \\
        Human (900 subset) & \ts{83.1}{94.8} & \ts{82.9}{86.7} & \ts{99.6}{97.5} & \ts{75.0}{96.7} & \ts{83.6}{95.0} & \ts{66.1}{93.3} & \ts{67.1}{93.3} & \ts{88.9}{92.5} & \ts{93.6}{100.0} & \ts{91.4}{98.3} \\
        Random & \ts{17.6}{--} & \ts{21.1}{--} & \ts{17.0}{--} & \ts{18.0}{--} & \ts{21.7}{--} & \ts{10.2}{--} & \ts{12.9}{--} & \ts{10.2}{--} & \ts{25.6}{--} & \ts{21.7}{--} \\
        Fixed-Time Oracle & \ts{24.2}{--} & \ts{37.5}{--} & \ts{25.9}{--} & \ts{35.0}{--} & \ts{28.8}{--} & \ts{17.1}{--} & \ts{14.2}{--} & \ts{9.3}{--} & \ts{22.5}{--} & \ts{27.2}{--} \\
        Blind VLM & \ts{2.8}{73.0} & \ts{1.7}{90.8} & \ts{0.0}{96.0} & \ts{6.4}{0.0} & \ts{8.4}{55.2} & \ts{0.0}{100.0} & \ts{0.0}{100.0} & \ts{8.7}{15.0} & \ts{0.0}{100.0} & \ts{0.0}{100.0} \\
        \midrule
        \rowcolor{rankviolet!30}[0pt][0pt]
        \multicolumn{11}{@{}l@{}}{\textcolor{black}{{Proprietary VLMs (API)}}} \\
        GPT-4o & \ts{24.5}{54.6} & \ts{73.8}{70.1} & \ts{45.6}{84.8} & \ts{0.6}{0.0} & \ts{3.8}{95.8} & \ts{8.5}{37.2} & \ts{7.9}{26.8} & \ts{26.7}{73.0} & \ts{25.8}{59.2} & \ts{27.9}{44.4} \\
        Gemini 3.1 Pro & \ts{26.4}{32.2} & \ts{71.2}{54.0} & \ts{67.1}{68.7} & \ts{5.1}{0.0} & \ts{23.6}{27.1} & \ts{10.3}{33.0} & \ts{17.6}{19.6} & \ts{11.7}{10.0} & \ts{25.1}{55.1} & \ts{6.1}{22.2} \\
        Claude Opus 5.5 & \ts{30.6}{45.0} & \ts{69.6}{63.2} & \ts{75.9}{79.8} & \ts{8.5}{10.0} & \ts{36.6}{46.9} & \ts{20.1}{61.7} & \ts{6.7}{10.3} & \ts{18.3}{21.0} & \ts{32.7}{79.6} & \ts{6.8}{32.3} \\
        \midrule
        \rowcolor{rankviolet!30}[0pt][0pt]
        \multicolumn{11}{@{}l@{}}{\textcolor{black}{{General-purpose VLMs}}} \\
        Qwen2.5-VL-7B-Instruct & \ts{4.5}{97.7} & \ts{31.7}{93.1} & \ts{0.0}{100.0} & \ts{1.5}{88.3} & \ts{0.0}{100.0} & \ts{0.0}{100.0} & \ts{0.0}{100.0} & \ts{0.0}{100.0} & \ts{7.6}{98.0} & \ts{0.0}{100.0} \\
        Qwen2.5-VL-32B-Instruct & \ts{11.8}{80.2} & \ts{59.6}{63.2} & \ts{2.2}{92.9} & \ts{0.4}{0.0} & \ts{1.6}{100.0} & \ts{0.0}{84.0} & \ts{0.0}{100.0} & \ts{5.3}{97.0} & \ts{37.1}{84.7} & \ts{0.0}{100.0} \\
        Qwen3-VL-8B-Instruct & \ts{17.7}{66.6} & \ts{70.4}{74.7} & \ts{25.9}{63.6} & \ts{2.6}{50.0} & \ts{3.0}{29.2} & \ts{6.4}{96.8} & \ts{8.6}{89.7} & \ts{11.0}{31.0} & \ts{26.9}{67.3} & \ts{4.8}{97.0} \\
        Qwen3-VL-32B-Instruct & \ts{26.8}{59.9} & \ts{74.6}{65.5} & \ts{60.1}{82.8} & \ts{2.1}{8.3} & \ts{23.8}{57.3} & \ts{12.4}{67.0} & \ts{13.1}{69.1} & \ts{21.0}{48.0} & \ts{26.9}{67.3} & \ts{7.5}{73.7} \\
        \midrule
        \rowcolor{rankviolet!30}[0pt][0pt]
        \multicolumn{11}{@{}l@{}}{\textcolor{black}{{Spatial / Embodied VLMs}}} \\
        Cosmos-Reason2-32B & \ts{21.6}{39.9} & \ts{73.3}{63.2} & \ts{48.7}{91.9} & \ts{0.0}{0.0} & \ts{1.0}{0.0} & \ts{12.4}{19.1} & \ts{18.0}{33.0} & \ts{10.3}{23.0} & \ts{19.6}{60.2} & \ts{10.9}{68.7} \\
        SenseNova-SI-1.1-8B & \ts{18.1}{85.6} & \ts{67.5}{73.6} & \ts{35.1}{69.7} & \ts{1.9}{73.3} & \ts{0.6}{94.8} & \ts{5.1}{97.9} & \ts{13.5}{80.4} & \ts{7.3}{95.0} & \ts{30.5}{86.7} & \ts{1.4}{99.0} \\
        Cambrian-S-7B & \ts{18.1}{44.1} & \ts{52.9}{90.8} & \ts{19.3}{52.5} & \ts{0.9}{16.7} & \ts{22.0}{34.4} & \ts{0.4}{1.1} & \ts{10.9}{39.2} & \ts{4.3}{3.0} & \ts{38.5}{72.4} & \ts{13.6}{86.9} \\
        VLM-3R & \ts{11.5}{28.3} & \ts{53.8}{43.7} & \ts{7.5}{18.2} & \ts{2.1}{23.3} & \ts{9.4}{10.4} & \ts{0.0}{0.0} & \ts{9.4}{61.9} & \ts{0.0}{1.0} & \ts{17.1}{35.7} & \ts{4.1}{60.6} \\
        \midrule
        \rowcolor{rankviolet!30}[0pt][0pt]
        \multicolumn{11}{@{}l@{}}{\textcolor{black}{{Streaming VLMs}}} \\
        TimeChat-Online-7B & \ts{16.6}{72.7} & \ts{76.2}{64.4} & \ts{30.3}{86.9} & \ts{0.0}{100.0} & \ts{0.0}{100.0} & \ts{0.0}{100.0} & \ts{0.0}{100.0} & \ts{4.0}{22.0} & \ts{17.8}{52.0} & \ts{21.1}{29.3} \\
        LiveCC-7B-Instruct & \ts{11.4}{76.2} & \ts{40.4}{96.6} & \ts{0.4}{92.9} & \ts{0.4}{95.0} & \ts{2.2}{69.8} & \ts{2.6}{81.9} & \ts{1.9}{89.7} & \ts{8.0}{45.0} & \ts{21.5}{62.2} & \ts{25.5}{52.5} \\
        Dispider & \ts{9.3}{48.4} & \ts{18.8}{72.4} & \ts{11.0}{60.6} & \ts{22.0}{6.7} & \ts{0.2}{100.0} & \ts{2.1}{0.0} & \ts{4.9}{21.6} & \ts{8.0}{15.0} & \ts{16.7}{59.2} & \ts{0.0}{100.0} \\
        \midrule
        \rowcolor{rankviolet!30}[0pt][0pt]
        \multicolumn{11}{@{}l@{}}{\textcolor{black}{{Streaming Spatial VLMs}}} \\
        Spatial-TTT-nano & \ts{6.2}{1.1} & \ts{20.4}{5.7} & \ts{9.2}{3.0} & \ts{8.3}{0.0} & \ts{15.6}{1.0} & \ts{0.0}{0.0} & \ts{0.0}{0.0} & \ts{0.0}{0.0} & \ts{2.2}{0.0} & \ts{0.3}{0.0} \\
        Stream3D-VLM-4B & \ts{4.8}{87.3} & \ts{28.8}{88.5} & \ts{9.2}{70.7} & \ts{0.6}{58.3} & \ts{0.4}{97.9} & \ts{1.3}{92.6} & \ts{1.5}{87.6} & \ts{0.0}{99.0} & \ts{0.7}{95.9} & \ts{0.7}{94.9} \\
        \midrule
        \rowcolor{rankviolet!30}[0pt][0pt]
        \multicolumn{11}{@{}l@{}}{\textcolor{black}{{+ ProSpace SFT (Prefix Polling)}}} \\
        Qwen3-VL-8B-Instruct & \ts{51.8}{70.8} & \ts{70.8}{65.5} & \ts{70.6}{76.8} & \ts{59.6}{85.0} & \ts{59.8}{61.5} & \ts{24.8}{90.4} & \ts{42.3}{63.9} & \ts{52.3}{63.0} & \ts{34.2}{52.0} & \ts{51.4}{78.8} \\
        VLM-3R & \ts{47.9}{68.0} & \ts{71.2}{56.3} & \ts{68.0}{62.6} & \ts{43.4}{73.3} & \ts{61.2}{82.3} & \ts{15.0}{66.0} & \ts{41.9}{75.3} & \ts{45.3}{64.0} & \ts{35.3}{48.0} & \ts{50.0}{83.8} \\
        Spatial-TTT-nano & \ts{41.2}{63.7} & \ts{77.1}{64.4} & \ts{63.2}{63.6} & \ts{11.3}{83.3} & \ts{47.2}{63.5} & \ts{9.8}{63.8} & \ts{47.6}{78.4} & \ts{40.3}{57.0} & \ts{34.9}{42.9} & \ts{39.5}{56.6} \\
        \midrule
        \rowcolor{rankviolet!30}[0pt][0pt]
        \multicolumn{11}{@{}l@{}}{\textcolor{black}{{+ ProSpace SFT (Native Streaming)}}} \\
        VideoLLM-online-8B & \ts{49.4}{71.9} & \ts{46.7}{32.2} & \ts{64.9}{69.7} & \ts{42.5}{76.7} & \ts{31.6}{41.7} & \ts{15.0}{97.9} & \ts{35.2}{77.3} & \ts{60.7}{88.0} & \ts{77.5}{85.7} & \ts{70.4}{77.8} \\
        Qwen2.5-VL-7B-Instruct & \ts{61.7}{78.7} & \ts{73.8}{56.3} & \ts{83.8}{88.9} & \ts{52.4}{90.0} & \ts{51.8}{56.2} & \ts{37.2}{86.2} & \ts{50.6}{82.5} & \ts{62.0}{82.0} & \ts{70.2}{75.5} & \ts{73.5}{90.9} \\
        Stream3D-VLM-8B & \ts{68.1}{84.8} & \ts{69.6}{57.5} & \ts{88.2}{88.9} & \ts{49.4}{88.3} & \ts{89.4}{99.0} & \ts{46.6}{90.4} & \ts{56.6}{94.8} & \ts{66.7}{81.0} & \ts{71.6}{69.4} & \ts{75.2}{93.9} \\
        \midrule
        \textbf{ProSpace-VLM} (w/ CUT3R) & \ts{71.8}{87.3} & \ts{70.8}{54.0} & \ts{89.0}{90.9} & \ts{58.3}{93.3} & \ts{83.4}{100.0} & \ts{69.2}{96.8} & \ts{61.4}{88.7} & \ts{67.3}{85.0} & \ts{72.4}{81.6} & \ts{74.1}{94.9} \\
        \textbf{ProSpace-VLM} (w/ StreamVGGT) & \ts{71.3}{85.5} & \ts{70.8}{47.1} & \ts{87.7}{91.9} & \ts{53.8}{83.3} & \ts{82.8}{99.0} & \ts{73.9}{94.7} & \ts{53.6}{89.7} & \ts{70.7}{88.0} & \ts{70.5}{80.6} & \ts{77.6}{94.9} \\
        \bottomrule
    \end{tabular}%
    }
    \label{tab:prospace_bench}
    \endgroup
\end{table*}

%% file: main/partial/fig_analysis_attention.tex
\begin{figure*}[t]
    \centering
    \includegraphics[width=0.9\textwidth]{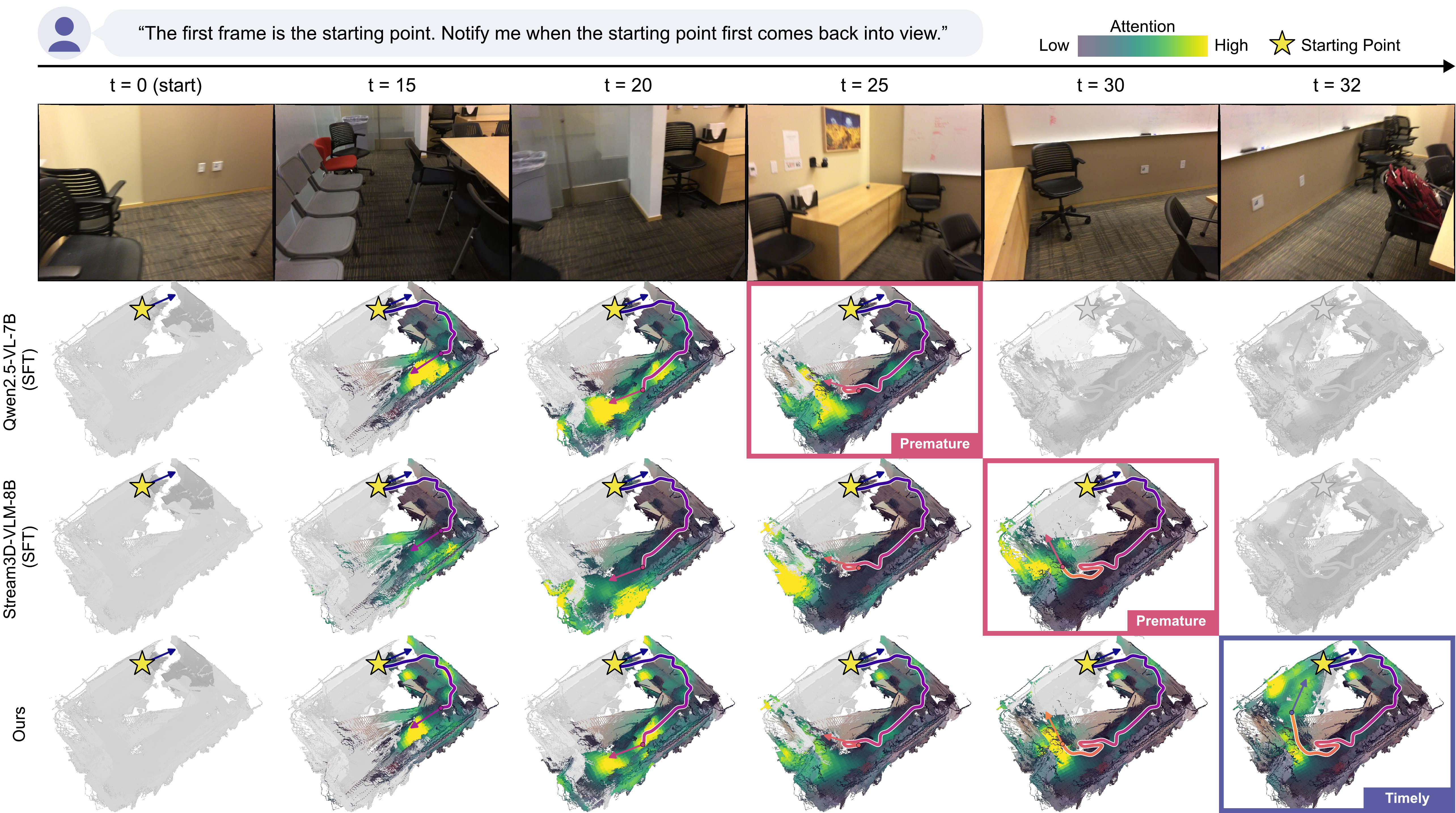}
    \caption{\textbf{Analysis of attention visualization.}
    Rows below the observed frames show the decision token's attention over cached visual tokens, placed at their 3D positions.
    Boxes mark each model's response frame.}
    \label{fig:analysis_attention}
    \vspace{-4mm}
\end{figure*}

%% file: main/partial/fig_analysis_bench_breakdown.tex
\begin{figure}[t]
    \centering
    \includegraphics[width=\columnwidth]{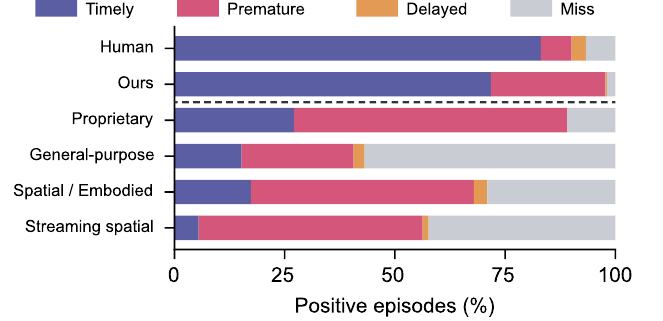}
    \caption{\textbf{Response timing breakdown.}
    Distribution of timely, premature, delayed, and missed responses.
    The dashed line separates human and ProSpace-VLM from zero-shot baselines.}
    \label{fig:analysis_bench_breakdown}
\end{figure}

%% file: main/partial/fig_analysis_failure_case.tex
\begin{figure}[t]
    \centering
    \includegraphics[width=\columnwidth]{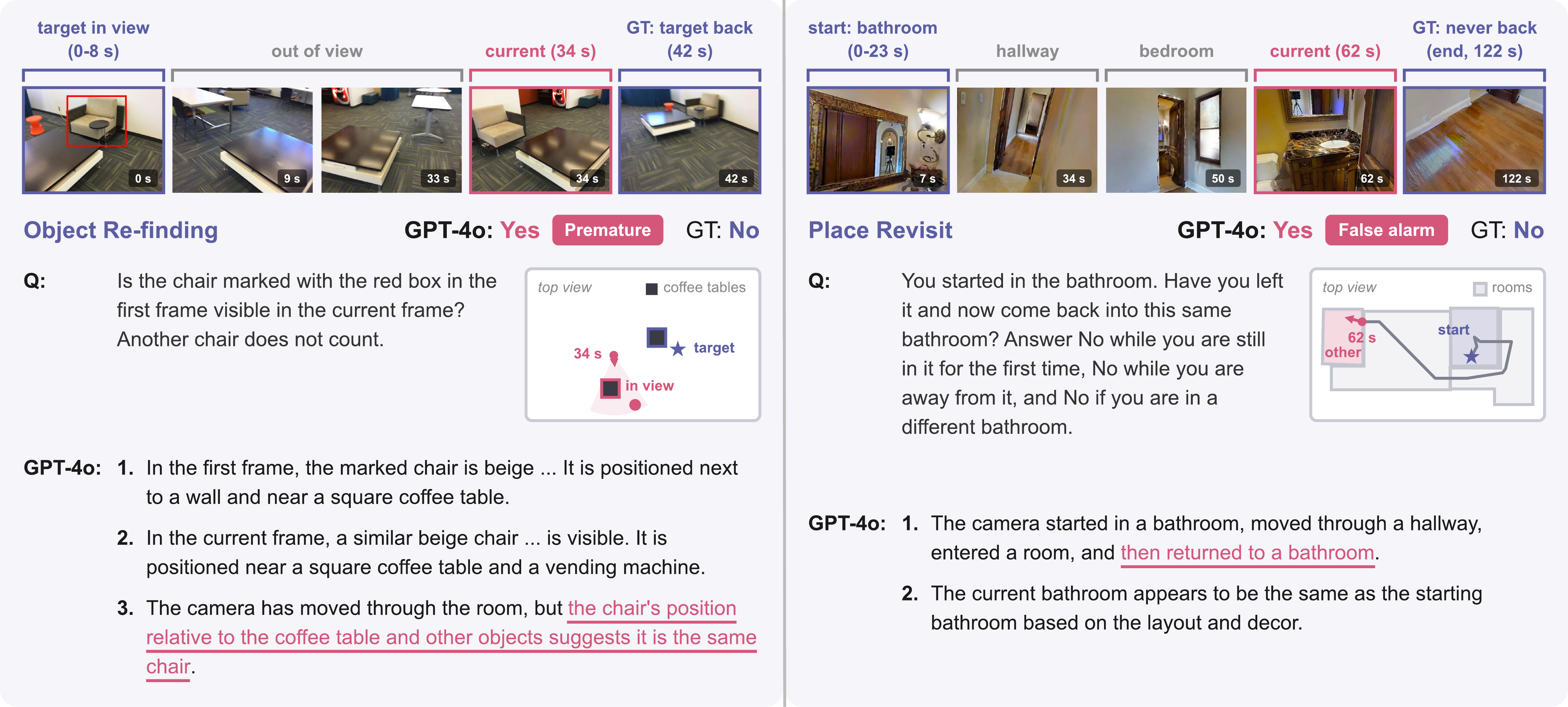}
    \caption{\textbf{Analysis of failure cases.}
    On \textit{Object Re-finding}, GPT-4o~\cite{hurst2024gpt4o} responds to a look-alike chair, citing its position relative to another identical coffee table.
    On \textit{Place Revisit}, it mistakes a different bathroom for the starting one.
    Both failures suggest reliance on appearance and local context, highlighting the importance of maintaining spatial states over time.
    Zoom in for details.
    }
    \label{fig:analysis_failure_case}
    \vspace{-4mm}
\end{figure}

%% file: main/table/table_ablation.tex
\begin{table}[t]
    \centering
    \begingroup
    \setlength{\tabcolsep}{5pt}
    \renewcommand{\arraystretch}{1.12}
    \footnotesize
    \caption{
        \textbf{Ablation study.} The first row is the fine-tuned Qwen2.5-VL-7B baseline.
        Combining all components yields the best performance.
    }
    \begin{tabular}{@{}ccccc@{}}
        \toprule
        \makecell{Camera\\Pose} &
        \makecell{Point\\Map} &
        \makecell{Multi-\\layer} &
        \makecell{Timely\\Response (\%)} &
        \makecell{Correct\\Abstention (\%)} \\
        \midrule
                        &                 &              & 61.7 & 78.7 \\
                        & $\checkmark$    & $\checkmark$ & 67.8 & 84.4 \\
        $\checkmark$    &                 & $\checkmark$ & 70.5 & 85.7 \\
        $\checkmark$    & $\checkmark$    &              & 70.9 & 85.9 \\
        $\checkmark$    & $\checkmark$    & $\checkmark$ & \textbf{71.8} & \textbf{87.3} \\
        \bottomrule
    \end{tabular}
    \vspace{-4mm}
    \label{tab:ablation}
    \endgroup
\end{table}

%% file: main/table/table_spatial_bench.tex
\begin{table}[!t]
    \centering
    \begingroup
    \caption{\textbf{Zero-shot evaluation on spatial benchmarks.}
    VSI\textsuperscript{$\dagger$} reports the mean over five question types.
    Avg. is the mean of the four benchmark scores.}
    \label{tab:spatial_bench}
    \setlength{\tabcolsep}{4pt}
    \renewcommand{\arraystretch}{1.1}
    \resizebox{\columnwidth}{!}{%
        \begin{tabular}{@{}lccccc@{}}
            \toprule
            Method &
            VSI\textsuperscript{$\dagger$}~\cite{yang2025thinkinginspace} &
            OST~\cite{lin2026ostbench} &
            UCS~\cite{wang2026keepitinmind} &
            OVO-S~\cite{li2026ovosbench} &
            Avg. \\
            \midrule
            Qwen2.5-VL-7B (SFT)
                & 31.0 & \textbf{40.5} & 38.5 & 40.6 & 37.7 \\
            Stream3D-VLM-8B (SFT)
                & 34.0 & 37.1 & 35.6 & 39.6 & 36.6 \\
            \textbf{Ours}
                & \textbf{34.9} & 39.7 & \textbf{39.3}
                & \textbf{42.4} & \textbf{39.1} \\
            \bottomrule
        \end{tabular}%
    }
    \vspace{-6mm}
    \endgroup
\end{table}

%% file: main/table/table_analysis_efficiency.tex
\begin{table}[t]
    \centering
    \caption{\textbf{Efficiency analysis.}
    FPS and memory are measured across 55 streams of 150 frames each.
    Mem.@1K denotes peak memory usage on 1,024-frame streams.}
    \label{tab:analysis_efficiency}
    \setlength{\tabcolsep}{4pt}
    \renewcommand{\arraystretch}{1.1}
    \resizebox{\columnwidth}{!}{%
        \begin{tabular}{@{}lrrrrc@{}}
            \toprule
            Method & FPS & Mem. (GB) & Mem.@1K (GB) & Timely / Abst. (\%) \\
            \midrule
            Qwen2.5-VL-7B (SFT)        & 10.1 & 18.2 & 33.4 & 61.7 / 78.7 \\
            Stream3D-VLM-8B            &  3.2 & 39.5 & OOM  & 68.1 / 84.8 \\
            Stream3D-VLM-8B (w/ GAVC)  &  2.4 & 38.5 & OOM  & 66.1 / 85.3 \\
            Ours                       &  5.8 & 21.2 & 36.4 & 71.8 / 87.3 \\
            Ours (w/ Spatial Token Selection)    &  6.7 & 19.3 & 19.4 & 70.8 / 87.9 \\
            \bottomrule
        \end{tabular}%
    }
\end{table}

%% file: main/partial/fig_analysis_stream_length.tex
\begin{figure}[t]
    \centering
    \includegraphics[width=\columnwidth]{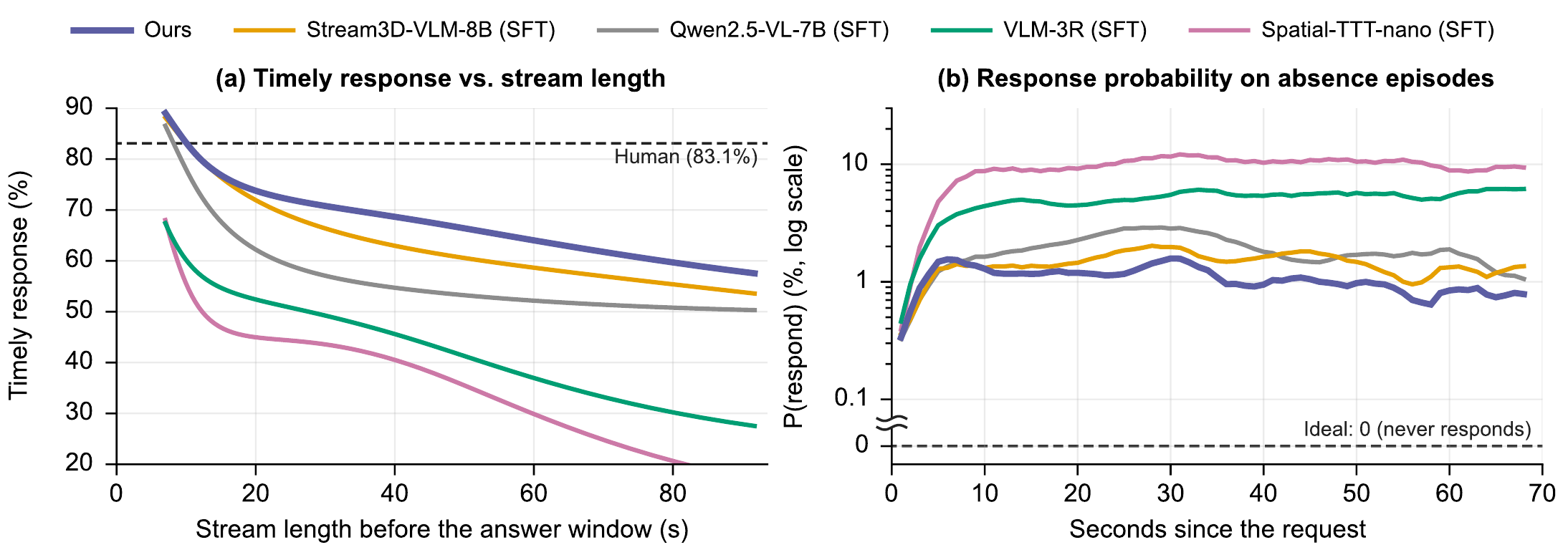}
    \caption{\textbf{Robustness to stream length and absence.}
    Left: timely response rate versus stream length before the answer window.
    Right: mean response probability per second on absence episodes.}
    \label{fig:analysis_stream_length}
    \vspace{-4mm}
\end{figure}

%% file: main/partial/fig_analysis_objperm.tex
\begin{figure}[t]
    \centering
    \includegraphics[width=\columnwidth]{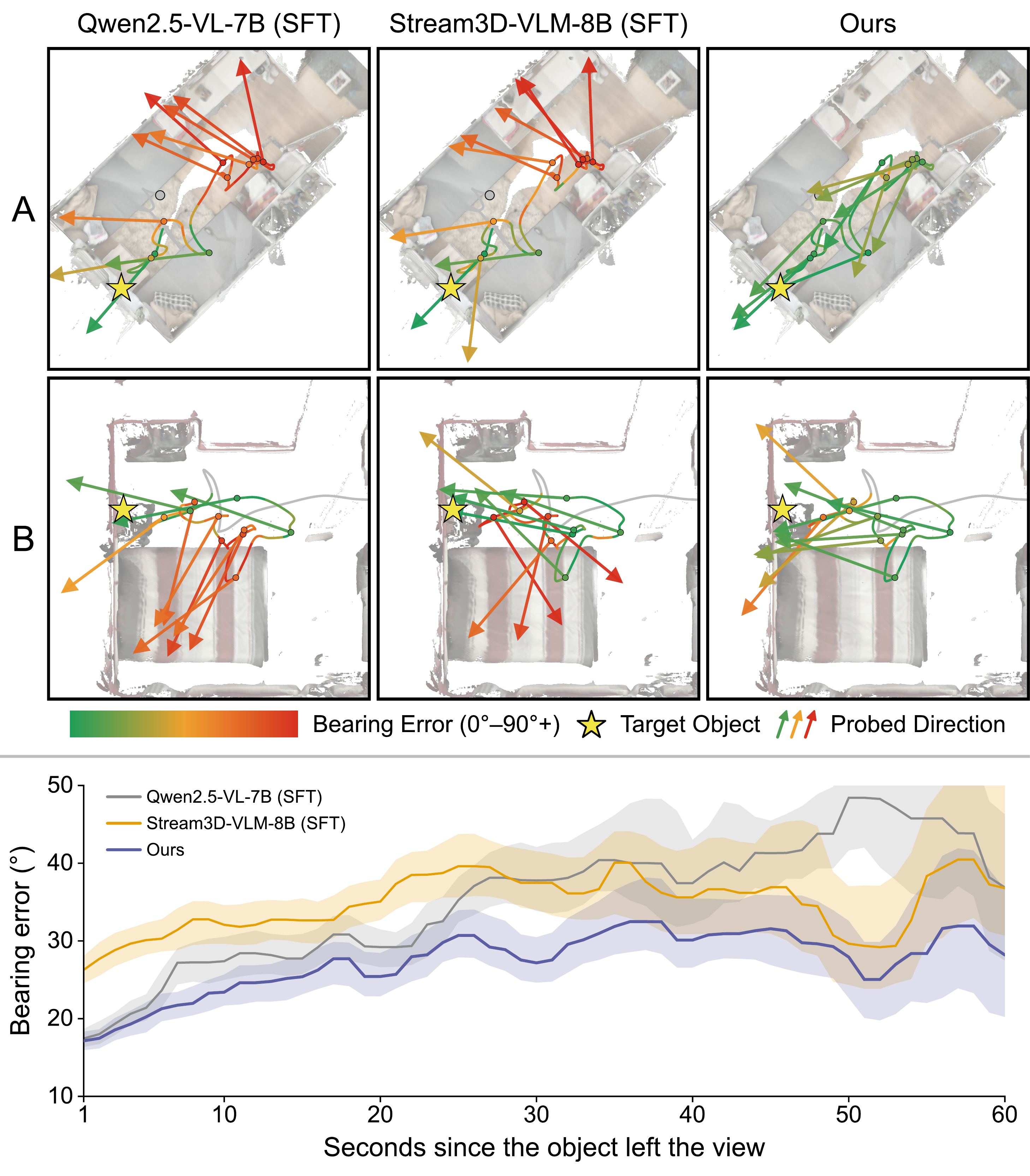}
    \vspace{-6mm}
    \caption{\textbf{Analysis of object permanence.}
    Top: target bearings decoded from decision-token states by a linear probe after the target leaves view.
    A and B denote different scenes.
    Bottom: bearing error vs.\ time out of view (median; shaded: 95\% bootstrap confidence interval over streams).}
    \label{fig:analysis_objperm}
    \vspace{-6mm}
\end{figure}

%% file: main/section/conclusion.tex
\section{Conclusion}
\label{sec:conclusion}
We study how models \textit{think in streams} through proactive spatial reasoning, where they respond when a requested condition holds or remain silent if it never occurs.
ProSpace-Bench evaluates this capability, revealing substantial gaps between VLMs and humans.
To support continual spatial decisions, we introduce ProSpace-VLM, which injects explicit geometric encodings into intermediate VLM layers to use the existing KV cache as spatial memory.
A limitation of ProSpace is its focus on response timing, without jointly evaluating higher-level reasoning or physical actions.
Moreover, rule-based episode construction may not capture the full diversity of spatial conditions encountered in real-world settings.